\documentclass[letterpaper, 10 pt, conference]{ieeeconf}  

\IEEEoverridecommandlockouts                              

\makeatletter
\let\NAT@parse\undefined
\makeatother
\usepackage[square,sort&compress]{natbib}
\newcommand{\bibfont}{\footnotesize}

\usepackage{balance} 

\usepackage{subcaption}

\usepackage{bm}
\usepackage{url}            
\usepackage{booktabs}       
\usepackage{amsfonts}       
\usepackage{nicefrac}       
\usepackage{microtype}      

\usepackage{afterpage}

\usepackage{graphicx}

\usepackage{epsfig}

\usepackage{xcolor}

\usepackage{times} 
\makeatletter
\@ifundefined{proof}{\usepackage{amsthm}}{}
\makeatother
\usepackage{amsmath}
\usepackage{amssymb}

\usepackage{subcaption}
\usepackage{caption}

\usepackage{bm}
\usepackage{xspace}

\allowdisplaybreaks

\usepackage{multirow}

\usepackage{makecell}

\usepackage{algorithm}
\usepackage{algorithmic}

\def\app#1#2{%
  \mathrel{%
    \setbox0=\hbox{$#1\sim$}%
    \setbox2=\hbox{%
      \rlap{\hbox{$#1\propto$}}%
      \lower1.1\ht0\box0%
    }%
    \raise0.25\ht2\box2%
  }%
}

\newcommand{\Kalmant}{\texttt{EnKTS}\xspace}
 
\def\p(#1|#2){p\left(#1 \, | \, #2\right)}

\def\BibTeX{{\rm B\kern-.05em{\sc i\kern-.025em b}\kern-.08em
    T\kern-.1667em\lower.7ex\hbox{E}\kern-.125emX}}
\title{\LARGE \bf
Bayesian Inferential Motion Planning Using Heavy-Tailed  Distributions
}

\author{Ali Vaziri, Iman Askari, and Huazhen Fang
\thanks{A. Vaziri, I. Askari, and H. Fang are with the
Department of Mechanical Engineering,
        University of Kansas, Lawrence, KS 66045, USA.
        Email: {\tt\small \{alivaziri,askari,fang\}@ku.edu}}%
}

\begin{document}

\maketitle
\thispagestyle{empty}
\pagestyle{empty}

\vspace{-100mm}

\begin{abstract}
Robots rely on motion planning to navigate safely and efficiently while performing various tasks. In this paper, we investigate motion planning through Bayesian inference, where motion plans are inferred  based on planning objectives and constraints. However, existing Bayesian motion planning methods often struggle to explore low-probability regions of the planning space, where high-quality plans may reside. To address this limitation, we propose the use of heavy-tailed distributions---specifically, Student's-$t$ distributions---to enhance probabilistic inferential search for motion plans. We develop a novel sequential single-pass smoothing approach that integrates Student's-$t$ distribution with Monte Carlo sampling. A special case of this approach is ensemble Kalman smoothing, which depends on short-tailed Gaussian distributions. We validate the proposed approach through simulations in autonomous vehicle motion planning, demonstrating its superior performance in planning, sampling efficiency, and constraint satisfaction compared to ensemble Kalman smoothing. While focused on motion planning, this work points to the broader potential of heavy-tailed distributions in enhancing probabilistic decision-making in robotics.

\end{abstract}

\section{Introduction}  \label{sec:Introduction}

Robotics is one of the most enabling technologies nowadays to transform myriad industry sectors and enhance human life. Robots' successful operation   relies on various intelligent decision-making capabilities. Among them,   motion planning is as foundational as it is flourishing, and the literature has presented many advances. 
Random sampling is a powerful tool for motion planning to search nonconvex, high-dimensional spaces.  Rapidly exploring random tree (RRT) is a popular method in this regard, due to its straightforward design, wide applicability, and almost sure  convergence to local optima~\cite{RRT}. It has inspired many variants,  including   kinodynamic RRT~\cite{frazzoli2002real} for dynamic constraints, closed-loop RRT~\cite{kuwata2009real}, and $\textrm{RRT}^*$ with optimality-aware design~\cite{karaman2011sampling, hwan2013optimal}.  Other important algorithms include Dijkstra's algorithm
~\cite{Dijkstra},  A*~\cite{A_star}, and potential  field algorithms~\cite{Potential_field, Potential_field_pathP}. These methods, however, make limited use of dynamic models. Optimization-driven methods hence have gained growing importance, with their capability of including predictive models and operation constraints to improve the generation of motion plans~\cite{Claussmann, thornton2016incorporating, eiras2021two, funke2016collision}. A diversity of techniques, e.g., iLQR and NMPC~\cite{chen2019autonomous, liu2017path, wei2022mpc}, have been studied in the literature. Yet, receding-horizon-constrained optimization at the core of these methods may require expensive computation, especially when facing highly nonlinear models or complex environments. Attempts to mitigate computation by introducing approximation will lead to performance loss.


 Bayesian inferential motion planning (BIMP) has recently attracted a growing interest due to its promise to address some complex problems. 
BIMP focuses on inferring the best motion plans using the planning objectives and constraints as the evidence within a Bayesian framework~\cite{berntorp2019motion, askari2025model}. It establishes a prior probability for motion plans, considers the likelihood of achieving the planning objectives and satisfying constraints given these plans, and then applies Bayes' rule to obtain the posterior probability of the plans. BIMP harnesses the power of probabilistic inference to offer multi-fold advantages. It allows to easily incorporate robot models into the predictive planning process and  enables the use of a plentiful number of nonlinear state estimation for motion planning. Especially, Monte Carlo sampling-based inference has gained significant use. For example, the studies in~\cite{berntorp2017path, askari2024motion} exploit particle filtering/smoothing and ensemble Kalman smoothing  to estimate  the   trajectories for   autonomous vehicles. What makes sampling-based BIMP even more appealing is its capability of tackling complicated  problems. For example, motion planning with machine learning dynamic models is challenging due to its complex optimization landscapes, but well-designed sampling schemes can achieve remarkably high computational efficiency~\cite{askari2025model, askari2024motion}. Further, the idea of Bayesian inference has shown effectiveness in dealing with various tricky problems in motion planning, e.g., probabilistic collision avoidance and uncertain or incomplete knowledge of the environment~\cite{patil2012estimating}. 

A key question for sampling-based BIMP is how to achieve {\em high-efficiency   search} in probabilistic state  spaces for {\em high-quality motion plans}. To address this question, the literature has introduced new sampling-based inference methods. In~\cite{askari2025model}, implicit particle filtering/smoothing seeks to concentrate sampling-based search within high-probability regions to improve both sampling efficiency and planning accuracy; in~\cite{askari2024motion}, a sequential single-pass ensemble Kalman smoothing is used to speed up computation. For sampling-based search, a main challenge is  that high-quality motion plans may lie in the less probable regions of the prior probability. Particle filtering/smoothing can partly handle the situation, yet demanding the use of many particles at more computational expenses.  The challenge is realistically  impactful for ensemble Kalman smoothing, as its dependence on Gaussian distribution assumptions would   hold it back from searching broadly. 

Heavy-tailed distributions present a promise to address the above question. These distributions are characterized by  non-exponentially bounded tails, which make a wide   spectrum of decisions   more probable in the motion planning context. If building upon this feature, probabilistic search can find out motion plans that look less probable in the prior but can increase the planning performance much. In this paper, we  explore advancing  BIMP with heavy-tailed distributions. Specifically, we consider Student's-$t$ distribution for the prior and likelihood in motion planning for autonomous vehicles and then in this setting, perform inference based on  Monte Carlo sampling. This development leads to a sequential smoothing algorithm for BIMP, which  takes a form similar to ensemble Kalman smoothing but shows significant improvement in both sample efficiency and planning performance for the challenging problem of BIMP with machine learning dynamic models for autonomous vehicles. We further reveal the connection of the proposed algorithm with motion planning based on NMPC.

\section{BIMP Problem Setup} \label{sec:Motion Planning Problem Formulation}

In this section, we set up the BIMP problem for autonomous vehicle motion planning. Consider an ego vehicle (EV) represented by a nonlinear dynamic model:
\begin{align}  \label{vehicle-model}
\bm x_{k+1} = f(\bm x_k, \bm u_k),
\end{align}
where $\bm x$ is the state including the position, heading angle, and velocity, and $\bm u$ is the input including the acceleration and steering angle. In this paper, we specifically use   a neural network dynamic model for $f(\cdot,\cdot)$ as in~\cite{askari2025model}, leveraging its data-driven learning capability to grasp vehicle dynamics.  This vehicle must move in such a way that $\bm x$ tracks a reference $\bm r$,   generated by a higher-level decision-making module, while its input $\bm u$ follows $\bm s$.  Some other so-called obstacle vehicles (OVs) are driving on the road, with their state denoted as $\bm x_k^{\mathrm{OV},i}$ for $i=1,2,\ldots, N_O$, where $N_O$ is the number of OVs. The EV must avoid colliding with the OVs, thus imposing the following constraint  in motion planning:
\begin{align}\label{collision-free-constraint}
d\left( \bm x_k,  \bm x_k^{\mathrm{OV},i} \right) \geq d_{\min}, \ \ i=1,2,\ldots,N_O,
\end{align}
where $d(\cdot,\cdot)$ is distance between the EV and OV $i$ with consideration of their geometry, and $d_{\min}$ is the minimum safe distance.  The vehicle should also stay within the lane boundaries. Its distance to the closes point $B$ on the boundaries, $d_B(\bm x_k)$, thus must satisfy
\begin{align}\label{lane-boundary-constraint}
d_B(\bm x_k) \leq L,
\end{align}
where $L$ is the lane width. Further, the control input is subject to 
\begin{align}\label{input-constraint}
\bm u_{\min} \leq \bm u_t \leq \bm u_{\max},
\end{align}
due to the vehicle's actuation limits and passenger comfort needs, with   $\bm u_{\min} $ and $ \bm u_{\max} $ denoting the lower and upper bounds of actuation. A passenger would also demand    the rate of change in actuation to be limited:
\begin{align}\label{incremental-input-constraint}
\Delta    \bm u_{\min} &\leq  \Delta \bm u_k \leq   \Delta \bm u_{\max}, \\
\Delta \bm u_k &= \bm u_k - \bm u_{k-1}
\end{align}
where $\Delta    \bm u_{\min}$ and $\Delta \bm u_{\max}$ are the lower and upper incremental actuation limits, respectively. 

Now, we aim to formulate a probabilistic motion planner, which aims to predict the   motion plans given the above driving setup. Loosely speaking,  the planner considers
\begin{align*}
\p ( \bm x_{k:k+H}, \bm u_{k:k+H}  |   \bm x \approx \bm r,  \bm u \approx \bm s,\eqref{vehicle-model}\textrm{-}\eqref{incremental-input-constraint} ),
\end{align*}
at time $k$ for the upcoming horizon $[k:k+H]$. To formalize this idea, we establish a virtual auxiliary system that describes the vehicle's future fictitious behavior  and observation:
\begin{align}\label{virtual-system}
\left\{
\begin{aligned} 
\bm x_{k+1} &=  f\left(\bm x_{k}, \bm u_{k} \right), \\
\bm u_{k+1} &=  \Delta u_{k+1} + \bm u_k, \\
\Delta \bm u_{k+1} &=  \bm w_{k}, \\
\bm r_k &= \bm x_{k} + \bm v_{\bm x, k} ,\\
\bm s_k &= \bm u_{k} + \bm v_{\bm u, k} ,\\
\bm z_k &= \bm \psi( \bm \phi (\bm x, \bm u, \Delta \bm u)) + \bm v_{\bm z, k} . 
\end{aligned}
\right.
\end{align} 
In above, $\bm x$, $\bm u$ and $\Delta \bm u$ together represent the state of this auxiliary system. The virtual observations of $\bm x$ and $\bm u$ are   $\bm r$ and $\bm s$, due to the anticipation that the vehicle will track the references. Further, all the constraints in~\eqref{collision-free-constraint}\textrm{-}\eqref{incremental-input-constraint} are summarized as 
\begin{align}\label{compact-constraint}
  \bm \phi (\bm x, \bm u, \Delta \bm u) \leq \bm 0.
\end{align}
We introduce $\bm z$ to measure   the constraint satisfaction by using a barrier function $\bm \psi(\cdot)$, which outputs zero if~\eqref{compact-constraint} and $\infty$ otherwise. Nominally, $\bm z=0$ as the vehicle is expected to adhere to   the constraints. Finally, $\bm v_{\bm x}$, $\bm v_{\bm u}$ and $\bm v_{\bm z}$ all random noises that reflects the uncertainty in the vehicle's future behavior. We can rewrite~\eqref{virtual-system} compactly as
\begin{subequations} \label{virtual-system-compact}
\begin{align}
\bar  {\bm x}_{k+1} &= \bar f\left(\bar {\bm x}_k \right) + \bar  {\bm w}_{k}, \label{virtual-system-dynamics}\\
\bar  {\bm y}_{k} &= \bar h\left( {\bar {\bm x}}_k \right)  + \bar  {\bm v}_{k}, \label{virtual-system-measurements}
\end{align}
\end{subequations}
where 
\begin{align*} 
\bar {\bm x}_k = 
\begin{bmatrix}
{\bm x}_k \cr
{\bm u}_k \cr
{\Delta \bm u}_k \cr
\end{bmatrix},  \ 
\bar {\bm y}_k = 
\begin{bmatrix}
\bm r_k  \cr 
\bm s_k  \cr
\bm z_k
\end{bmatrix}, \
\bar {\bm w}_k = 
\begin{bmatrix}
\bm 0 \cr
{\bm w}_k \cr
{\bm w}_k \cr
\end{bmatrix}, \
\bar {\bm v}_t = 
\begin{bmatrix}
\bm v_{\bm x, k} \cr  
\bm v_{\bm u, k} \cr
\bm v_{\bm z, k}
\end{bmatrix},
\end{align*}
and  $\bar f, \ \bar h$ are evident from the context. 

Given~\eqref{virtual-system-compact}, the probabilistic motion planner, or BIMP framework, seeks to address the following receding-horizon maximum a posterior estimation problem: 
\begin{equation} \label{MAP-for-virtual-system}
\bar{\bm{x}}^*_{k:k+H} = \arg\max_{\bar{\bm{x}}_{k:k+H}} \log \ \p(\bar{\bm{x}}_{k:k+H} | \bar{\bm{y}}_{k:k+H}, \bar{\bm{x}}_{k-1}).
\end{equation}
By design, this framework attempts to infer the motion plans that have the maximum posterior probability conditioned on  the driving objectives and constraints. What is shown in~\eqref{MAP-for-virtual-system}
is a state smoothing problem, which is has analytical solution for the nonlinear  $\bar f(\cdot)$. In the literature, this problem has been addressed using particle filtering/smoothing and ensemble Kalman smoothing~\cite{askari2024motion, berntorp2019motion}. Next, we will develop a new method that builds upon heavy-tailed distributions.

\section{BIMP Using Student's-$t$ Distributions}

The problem in~\eqref{MAP-for-virtual-system} mandates probabilistic search for the best motion plans. From a Bayesian perspective, the search entails evaluating 
\begin{align*} 
\p(\bar{\bm{x}}_{k:k+H} | \bar{\bm{y}}_{k:k+H}, \bar{\bm{x}}_{k-1})
& \propto 
\underbrace{\p(\bar{\bm{y}}_{k:k+H} | \bar{\bm{x}}_{k:k+H})}_{\mathrm{likelihood}} \\ & \quad\quad\quad \cdot \underbrace{p( \bar{\bm{x}}_{k:k+H}| \bar{\bm{x}}_{k-1})}_{\mathrm{prior}}.
\end{align*}
Here, $\p(\bar{\bm{x}}_{k:k+H} | \bar{\bm{y}}_{k:k+H})$ will present an extremely complex probability distribution landscape  due to the highly nonlinear neural network-based dynamics. As such, a motion plan with a low prior probability may have high likelihood. Such a motion plan may never be visited by the probabilistic search if using short-tailed distributions, e.g., Gaussian distribution. Heavy-tailed distributions come as a sensible alternative, by giving   less likely regions within the  probabilistic landscape higher probability. Therefore, they can help expand the reach, as well as the sampling efficiency, of the probabilistic search to find out better motion plans and enhance constraint satisfaction. Here, we consider Student's-$t$ distribution. They are not only heavy-tailed, but also enable closed-form update, making them   amenable to computation. 

Applying Bayes' rule to~\eqref{MAP-for-virtual-system}, we have
\begin{align}\label{single-pass-smoothing}\nonumber
\p(\bar {\bm x}_{k:t} | \bar {\bm y}_{k:t}, \bar{\bm{x}}_{k-1} ) 
&\propto  \p( \bar {\bm y}_t | \bar {\bm x}_t) \p( \bar {\bm x}_t | \bar {\bm x}_{t-1}) \\
&\quad\quad \cdot \p(\bar {\bm x}_{k:t-1} | \bar {\bm y}_{k:t-1} , \bar{\bm{x}}_{k-1}),
\end{align}
which  offers a sequential update from $\p(\bar {\bm x}_{k:t-1} | \bar {\bm y}_{k:t-1}, \bar{\bm{x}}_{k-1} )$ to  $\p(\bar {\bm x}_{k:t} | \bar {\bm y}_{k:t}, \bar{\bm{x}}_{k-1} )$ for $t=k, \ldots, k+H$. 
Let us define $\bm {\mathcal X}_k = \bm {\bar x}_{k:t} $ and $\bm {\mathcal Y}_k = \bm {\bar y}_{k:t} $ for notational simplicity. Going further, we propose to approximate $\p(\bm {\mathcal X}_t, \bm {\bar y}_t | \bm {\mathcal Y}_{t-1})$ as a Student's-$t$ distribution: 
\begin{align} \label{student_t-distribution-assumption}
\left.
\begin{bmatrix}
\bm {\mathcal X}_t \cr  \bm {\bar y}_t
\end{bmatrix} \; \right| \; \bm {\mathcal Y}_{t-1}
\sim
St\left(
\begin{bmatrix}
\widehat{\bm {\mathcal X}}_{t | t-1}  \cr  
\widehat{\bm{\bar y}}_{t|t-1}
\end{bmatrix},
\begin{bmatrix}
\bm P_{t|t-1}^{\bm {\mathcal X}} 
& \bm P_{t|t-1}^{\bm {\mathcal X \bm{\bar y}}} 
\cr
\left( \bm P_{t|t-1}^{\bm {\mathcal X \bm{\bar y}}} \right)^\top &
\bm P_{t|t-1}^{\bm{\bar  y}}
\end{bmatrix}, \nu
\right)
\end{align}
where $\widehat{\bm {\mathcal X}}$ and $\widehat{\bm{\bar y}}$ are the means of $\bm {\mathcal X}$ and $\bm{\bar y}$, and $\bm P$ are covariance matrices. For the considered problem, the Student's-$t$-based approximation is more realistic and suitable, as argued above. 

Given~\eqref{single-pass-smoothing}-\eqref{student_t-distribution-assumption}, a closed-form update is available to obtain the posterior probability $\p(\bm {\mathcal X}_t  | \bm {\mathcal Y}_{t}) $: 
\begin{subequations} \label{Kalman-update}
\begin{align} 
& \p(\bm {\mathcal X}_t  | \bm {\mathcal Y}_{t}) \sim St \left( \widehat{\bm {\mathcal X}}_{t | t}, \bm \Sigma_{t|t}^{\bm {\mathcal X}}, \nu^\prime \right ), \label{marginal_student_t_update}\\
&\widehat{\bm {\mathcal X}}_{t | t} = \widehat{\bm {\mathcal X}}_{t | t-1}  +  \bm {K}_t 
\left( \bar {\bm y}_t - \widehat{ \bar {\bm y}}_{t|t-1}  \right),\\ \label{Kalman-update-cov}
&\bm \Sigma_{t|t}^{\bm {\mathcal X}} = \frac{\nu + \delta }{\nu + n}(\bm P_{t|t-1}^{\bm {\mathcal X}} - \bm {K}_t \bm P_{t|t-1}^{\bar{\bm  y}} \bm {K}^\top_t), \\
&\bm {K}_t  = \bm P_{t|t-1}^{\bm {\mathcal X \bar{\bm y}}}  \left( \bm P_{t|t-1}^{\bar{\bm  y}} \right)^{-1}, \\
&\delta =  \left( \bar {\bm y}_t - \widehat{ \bar {\bm y}}_{t|t-1}  \right)^\top \left( \bm P_{t|t-1}^{\bar{\bm  y}} \right)^{-1} \left( \bar {\bm y}_t - \widehat{ \bar {\bm y}}_{t|t-1}  \right) ,\\
&\nu^\prime = \nu + n.
\end{align}
\end{subequations}
where $n$ is the dimension of the virtual measurements. Additionally, we assume 
\begin{align}\label{Student-t-approximation}
   {\bm w} \sim St \left(\bm 0; \bm \Sigma_{{ \bm w}}, \nu_{{\bm w}} \right), \    \bm v_{\bm x} \sim St \left(\bm 0; \bm \Sigma_{\bm v_{\bm x}}, \nu_{\bm x} \right),  \\
     \bm v_{\bm u} \sim St \left(\bm 0; \bm \Sigma_{\bm v_{\bm u}}, \nu_{\bm u} \right), \ \bm v_{\bm z} \sim St \left(\bm 0; \bm \Sigma_{\bm v_{\bm z}}, \nu_{\bm z} \right). 
\end{align}

To run the above Student's-$t$ update, we use Monte Carlo sampling to weakly approximate the involved probability distributions. 
At time $t-1$, we approximate  $\p(\bm {\mathcal X}_{t-1} | \bm {\mathcal Y}_{t-1})$ by an ensemble of samples, $\left\{\bar {\bm x}^j_{t-1|t-1}, j=1,2,\ldots,N\right\}$. Passing them through~\eqref{virtual-system-dynamics}, we can obtain 
\begin{align} \label{ensemble-mean-prediction}
\bar {\bm x}^j_{t|t-1} &= \bar f\left(\bar {\bm x}^j_{t-1|t-1} \right) + \bar  {\bm w}^j_{t-1}, \quad  j=1,2,\ldots, N,
\end{align} 
where $\bar  {\bm w}^j_{t-1}$ are   drawn from $St \left(\bm 0; \bm \Sigma_{\bm w}, \nu_{\bm w}\right)$. We join $\bar {\bm x}^j_{t|t-1}$ to  $\bm {\mathcal X}^j_{t-1|t-1}$ via the concatenating operation
\begin{align*}
\bm {\mathcal X}^j_{t|t-1}\xleftarrow{\mathrm{Concatenate}}{ \left(\bm {\mathcal X}^j_{t-1|t-1}, \bar {\bm x}^j_{t|t-1} \right)}. 
\end{align*}
We   then calculate the sample mean and covariance as 
\begin{subequations} 
\begin{align}
\widehat{\bm {\mathcal X}}_{t|t-1} 
&= \frac{1}{N}\sum_{j=1}^N \bm {\mathcal X}^j_{t|t-1}, \label{ensemble-sample-mean-prediction}\\
\bm P_{t|t-1}^{\bm {\mathcal X}}  
&= \frac{1}{N}\frac{\nu  - 2}{\nu} \sum_{j=1}^N \widetilde{{\bm {\mathcal X}}}^j_{t|t-1} \widetilde{{\bm {\mathcal X}}}^{j \ \top}_{t|t-1} \label{ensemble-sample-cov-prediction},
\end{align}
\end{subequations}
where $\widetilde{{\bm {\mathcal X}}} = {\bm {\mathcal X}} - \widehat{{\bm {\mathcal X}}}$.
We can now pass ${\bm {\bar x}}^j_{t|t-1}$    through~\eqref{virtual-system-measurements} to obtain the samples approximating $\p(  \bm {\bar y}_t | \bm {\mathcal Y}_{t-1})$: 
\begin{align} \label{ybar-sample}
\bm {\bar y}^j_{t|t-1} = \bm {\bar x}^j_{t|t-1} + \bar  {\bm v}^j_{t}, \quad  j=1,2,\ldots, N, 
\end{align}

The sample mean and covariances $\hat{\bm {\bar y}}_{t|t-1},\bm P_{t|t-1}^{\bm {\bar y}}, \bm P_{t|t-1}^{\bm {\mathcal X} \bm {\bar y}}$  are 
\begin{subequations}
\begin{align}
\hat{\bm {\bar y}}_{t|t-1} &= \frac{1}{N} \sum_{j=1}^N \bm {\bar y}^j_{t|t-1}, \label{ybar-sample-mean}\\
\bm P_{t|t-1}^{\bm {\bar y}} 
&= \frac{1}{N}\frac{\nu - 2}{\nu} \sum_{j=1}^N \tilde{\bm {\bar y}}^j_{t|t-1} \tilde{\bm {\bar y}}^{j, \ \top}_{t|t-1}, \label{P-ybar-update} \\
\bm P_{t|t-1}^{\bm {\mathcal X} \bm {\bar y}} 
&= \frac{1}{N}\frac{\nu - 2}{\nu} \sum_{j=1}^N \widetilde{{\bm {\mathcal X}}}^j_{t|t-1} \tilde{\bm {\bar y}}_{t|t-1}^{j, \ \top}, \label{P-Xybar-update}
\end{align}
\end{subequations}
where  $ \tilde{\bm {\bar y}} =  \bm {\bar y} - \widehat{\bm {\bar y}}$.
Finally, we can use~\eqref{Kalman-update} to update the state samples as
\begin{subequations}
\begin{align}
 &{\bm {\mathcal X}}^j_{t | t} = {\bm {\mathcal X}}^j_{t | t-1}  +  \bm {K}_t 
\left( \bar {\bm y}^j_t - \widehat{ \bar {\bm y}}_{t|t-1}  \right), \label{Kalmant-update-mean}  \\
&\widehat{\bm {\mathcal X}}_{t|t} = \frac{1}{N} \sum_{j=1}^N {{\bm {\mathcal X}}}^j_{t|t}, \label{Kalmant-update-sample-mean} \\
&\bm {K}_t  = \bm P_{t|t-1}^{\bm {\mathcal X \bar{\bm y}}}  \left( \bm P_{t|t-1}^{\bar{\bm  y}} \right)^{-1}, \label{Kalmant-update-gain} \\
&\delta =  \frac{1}{N} \sum_{j=1}^N\left( \bar {\bm y}^j_t - \widehat{ \bar {\bm y}}_{t|t-1}  \right)^\top \left( \bm P_{t|t-1}^{\bar{\bm  y}} \right)^{-1} \left( \bar {\bm y}^j_t - \widehat{ \bar {\bm y}}_{t|t-1}  \right), \label{Kalmant-update-Delta}\\
&\bm \Sigma_{t|t}^{\bm {\mathcal X}} = \frac{\nu + \delta }{\nu + n}(\bm P_{t|t-1}^{\bm {\mathcal X}} - \bm {K}_t \bm P_{t|t-1}^{\bar{\bm  y}} \bm {K}^\top_t), \label{Kalmant-update-cov} \\
&\nu' = \nu + n. \label{Kalmant-update-nu}
\end{align}
\end{subequations}

Running the above procedure sequentially over the horizon $t=k,\ldots,k+H$, we obtain a single-pass smoother to solve~\eqref{MAP-for-virtual-system}. This smoother  bears resemblance to the ensemble Kalman smoother in   form; the latter is indeed a special case of the former when $\nu\rightarrow\infty$.  However, the underlying Student's-$t$ distribution    will empower the proposed smoother to achieve more accurate and sample-efficient smoothing performance   for motion planning, as will be shown in Section~\ref{sec:Connection to Nonlinear Model Predictive Control}.  We name the algorithm as \Kalmant, summarizing it in Algorithm~\ref{alg:EnKTS}.

\begin{algorithm}[t]
\caption{\Kalmant Bayesian motion planner: Heavy-tailed single pass smoother}
\label{alg:EnKTS}
\begin{algorithmic}[1]

\STATE Set up the Bayesian inference problem in~\eqref{virtual-system}
\STATE Set up the virtual system in~\eqref{virtual-system-compact}
 
\FOR{$k=1, 2, \ldots$}

\STATE Initialize  $ \bar {\bm x}_{k|k}^j$ for $j = 1, \ldots, N $ 

        \FOR{$t=k+1, \ldots, k+H$}

\vspace{5pt}

\item[]{ \color{gray} \textsf{// Prediction}  }

\STATE Compute  $  \bar {\bm x}_{t|t-1}^i$ and via~\eqref{ensemble-mean-prediction}

\STATE $\bm {\mathcal X}^j_{t|t-1}\xleftarrow{\mathrm{Concatenate}}{(\bm {\mathcal X}^j_{t-1|t-1}, \bar {\bm x}^j_{t|t-1})}$

\STATE Compute  $ \widehat{\bm {\mathcal X}}_{t|t-1}$ via~\eqref{ensemble-sample-mean-prediction} 

\STATE{\color{gray} Compute  $ \bm P_{t|t-1}^{\bm {\mathcal X}}$ via~\eqref{ensemble-sample-cov-prediction} \quad\quad  \% Skippable} 

\vspace{5pt}

\item[] { \color{gray}  \textsf{//   Update}}  

\STATE Compute $\bar {\bm y}_{t|t-1}^j$, $\hat{\bar {\bm y}}_{t|t-1}$, $\bm P_{t|t-1}^{\bar {\bm y}}$, and $\bm P_{t|t-1}^{\bm {\mathcal X}  \bar {\bm y}}$ via~\eqref{ybar-sample}-\eqref{ybar-sample-mean}
-\eqref{P-ybar-update}-\eqref{P-Xybar-update}

\STATE Compute $ {\bm {\mathcal X}}_{t|t}^i$ and $\widehat{\bm {\mathcal X}}_{t|t}$ via~\eqref{Kalmant-update-mean}-\eqref{Kalmant-update-sample-mean}

\STATE  {\color{gray} Compute $\bm \Sigma_{t|t}^{\bm {\mathcal X}}$ via~\eqref{Kalmant-update-cov} \quad\quad  \% Skippable}

\STATE Update the degree of freedom parameter as \eqref{Kalmant-update-nu}

\ENDFOR

\STATE Extract $\widehat {\bar {\bm x}}_{k|k:k+H}$  from $\widehat {\bm {\mathcal X}}_{k+H|k+H}$, and then $\widehat {\bm u}_{k|k:k+H}$ from $\widehat {\bar {\bm x}}_{k|k:k+H}$ 

\STATE Apply $\widehat {\bm u}_{k|k:k+H}$  to the dynamical system and go to the next time step $k+1$

\ENDFOR
 
\end{algorithmic}
\end{algorithm}

\section{Connection to Nonlinear Model Predictive Control} \label{sec:Connection to Nonlinear Model Predictive Control}

In this section, we investigate  the  connections of BIMP with Student's-$t$ distributions with NMPC-based motion planning. 

Consider the problem in~\eqref{MAP-for-virtual-system}. 
By Bayes' rule and the Markovian property of~\eqref{virtual-system-compact}, we have
\begin{align*}
\p(\bar {\bm x}_{k:k+H}| \bar {\bm y}_{k:k+H}, \bar{\bm{x}}_{k-1}) &\propto \prod_{t=k}^{k+H} \p(\bar {\bm y}_t | \bar {\bm x}_t)  \p(\bar {\bm x}_t | \bar {\bm x}_{t-1}) .
\end{align*}
The log-likelihood then is given by 
\begin{align*}
&\log \p(\bar {\bm x}_{k:k+H}| \bar {\bm y}_{k:k+H}, {\bm x}_{k-1}) \propto 
\sum_{t=k}^{k+H} \log \p( \bm r_t | \bm x_t) \\
&\quad \quad \quad \quad + \sum_{t=k}^{k+H} \log \p( \bm s_t | \bm u_t) 
+\sum_{t=k}^{k+H} \log p\left(  \Delta \bm u_t\right),
\end{align*}
because $ \p(\bar {\bm y}_t | \bar {\bm x}_t )  = \p( \bm r_t | \bm x_t)  \p( \bm s | \bm u_t)$, and $\p(\bar {\bm x}_t | \bar {\bm x}_{t-1}) = p\left(  \Delta \bm u_t \right)$. Given~\eqref{virtual-system}, $\bm r_t \; | \; \bm x_t \sim St \left( \bm x_t; \bm \Sigma_{\bm v_{\bm x}}, \nu_{\bm x} \right)$, $\bm s_t \; | \; \bm u_t \sim St \left( \bm u_t;   \bm \Sigma_{\bm v_{\bm u}}, \nu_{\bm u} \right)$, and $  \Delta \bm u_t \sim St \left( \bm 0;   \bm \Sigma_{\bm w}, \nu_{\bm w} \right)$. By~\eqref{Student-t-approximation}, it follows that
\begin{align*}
& \log \p(\bar {\bm x}_{k:k+H}| \bar {\bm y}_{k:k+H}, {\bm x}_{k-1}) \\ 
& \propto \sum_{t=k}^{k+H} (-\frac{\nu_{\bm x} + n_{\bm x}}{2})\log \Big[1+\frac{1}{\nu_{\bm x}}(\bm{x}_t - \bm r_t)^\top \bm \Sigma_{\bm v_{\bm x}}^{-1} (\bm{x}_t - \bm r_t) \Big] \\ 
& + \sum_{t=k}^{k+H} (-\frac{\nu_{\bm u} + n_{\bm u}}{2})\log \Big[1+\frac{1}{\nu_{\bm u}}(\bm{u}_t - \bm s_t)^\top \bm \Sigma_{\bm v_{\bm u}}^{-1} (\bm{u}_t - \bm s_t) \Big]\\ 
& +\sum_{t=k}^{k+H} (-\frac{\nu_{\Delta \bm u} + n_{ \Delta \bm u}}{2}) \log \Big[1+\frac{1}{\nu_{ \Delta \bm u}}  \Delta \bm{u}_{t}^\top \bm \Sigma_{\bm w}^{-1}  \Delta \bm{u}_{t} \Big].
\end{align*}
This indicates that BIMP approximately solves an NMPC problem carrying the above non-quadratic cost function. Further, if $\nu_{\bm x}$, $\nu_{\bm u}$ and $\nu_{\Delta \bm u}\rightarrow \infty$, the involved Student's-$t$ distributions   will reduce to be Gaussian, and the above log-likelihood will lead to a regular-form quadratic cost function. In this case, BIMP solves an NMPC motion planning problem in the form of  
\begin{subequations} \label{optimal_control_problem}
\begin{align}
\min \quad &  J\left(\bm x_{k:k+H}, \bm u_{k:k+H},  \Delta \bm u_{k:k+H}\right) \\
\text{s.t.} \quad &  \bm x_{t +1} = f(\bm x_t, \bm u_t),\quad t=k,\dots,k+H,\\
            \quad &  \bm \phi(\bm x_t, \bm u_t, \Delta \bm u_t)\leq 0,\quad t=k,\dots,k+H, \\
            &\Delta \bm u_{t+1} = \bm u_{t} - \bm u_{t+1},\quad t=k-1,\dots,k+H-1.
\end{align}
\end{subequations}
where 
\begin{align*} \label{tracking_cost}
J\left(\bar{\bm{x}}_{k:k+H}\right) &=  \sum_{t=k}^{k+H} 
\Big[ (\bm{x}_t - \bm r_t)^\top \bm \Sigma_{\bm v_{\bm x}}^{-1} (\bm{x}_t - \bm r_t) \\ 
&+ (\bm{u}_t - \bm s_t)^\top \bm \Sigma_{\bm v_{\bm u}}^{-1} (\bm{u}_t - \bm s_t)   
 +  \Delta \bm{u}_{t}^\top \bm \Sigma_{\bm w}^{-1}  \Delta \bm{u}_{t} \Big]. 
\end{align*}

This analysis shows that BIMP conveys optimality in a certain form in motion planning. As another interesting implication, BIMP can offer an alternative way to solve NMPC problems---in this regard, the \Kalmant algorithm is a viable solver. Compared to   gradient-based optimization solvers, the \Kalmant algorithm will have stronger performance in handling complex nonlinear dynamics, e.g., a neural network dynamic model, drawing from our prior study in~\cite{askari2024motion}. This is due to its capability of performing sequential estimation and sampling-based search for optimal  control actions.

\section{Simulation Results} \label{sec:Simulation Results}
In this section, we apply the \Kalmant algorithm to motion planning of an autonomous vehicle. Following~\cite{askari2024motion}, we trained a feedforward neural network dynamic model for the EV vehicle with two hidden layers each with $128$ nodes. We use EnKS in~\cite{askari2024motion} to benchmark the efficacy of the proposed method in handling challenging overtaking and emergency braking scenarios with tight constraints. In the simulations, we used 50 particles and set $H=20$ with sampling time $\Delta t=0.1$ seconds with all of the other parameters the same for both methods. Our simulation results show that \Kalmant outperforms EnKS and is able to generate safe motion plans.

\begin{figure*}[t]
    \centering
    \begin{minipage}{0.47\textwidth}
        \centering
        \includegraphics[trim={5cm 5cm 4.8cm 2cm}, clip, width=\textwidth]{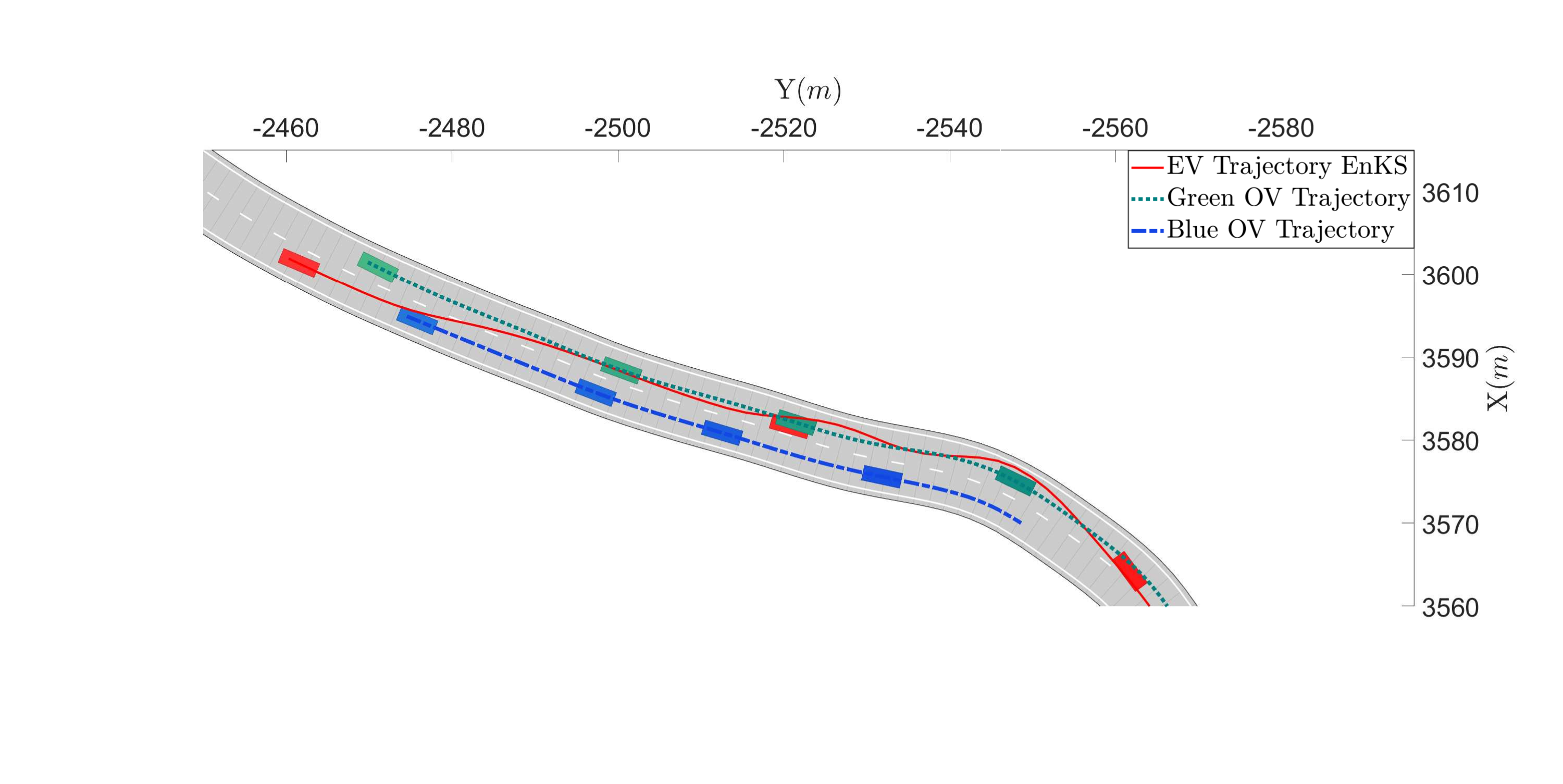}
        \subcaption{} \label{fig:Overtaking_Traj_EnKS}
    \end{minipage}%
    \hfill
    \begin{minipage}{0.53\textwidth}
        \centering
        \includegraphics[trim={4cm 5cm 0cm 2cm}, clip, width=\textwidth]{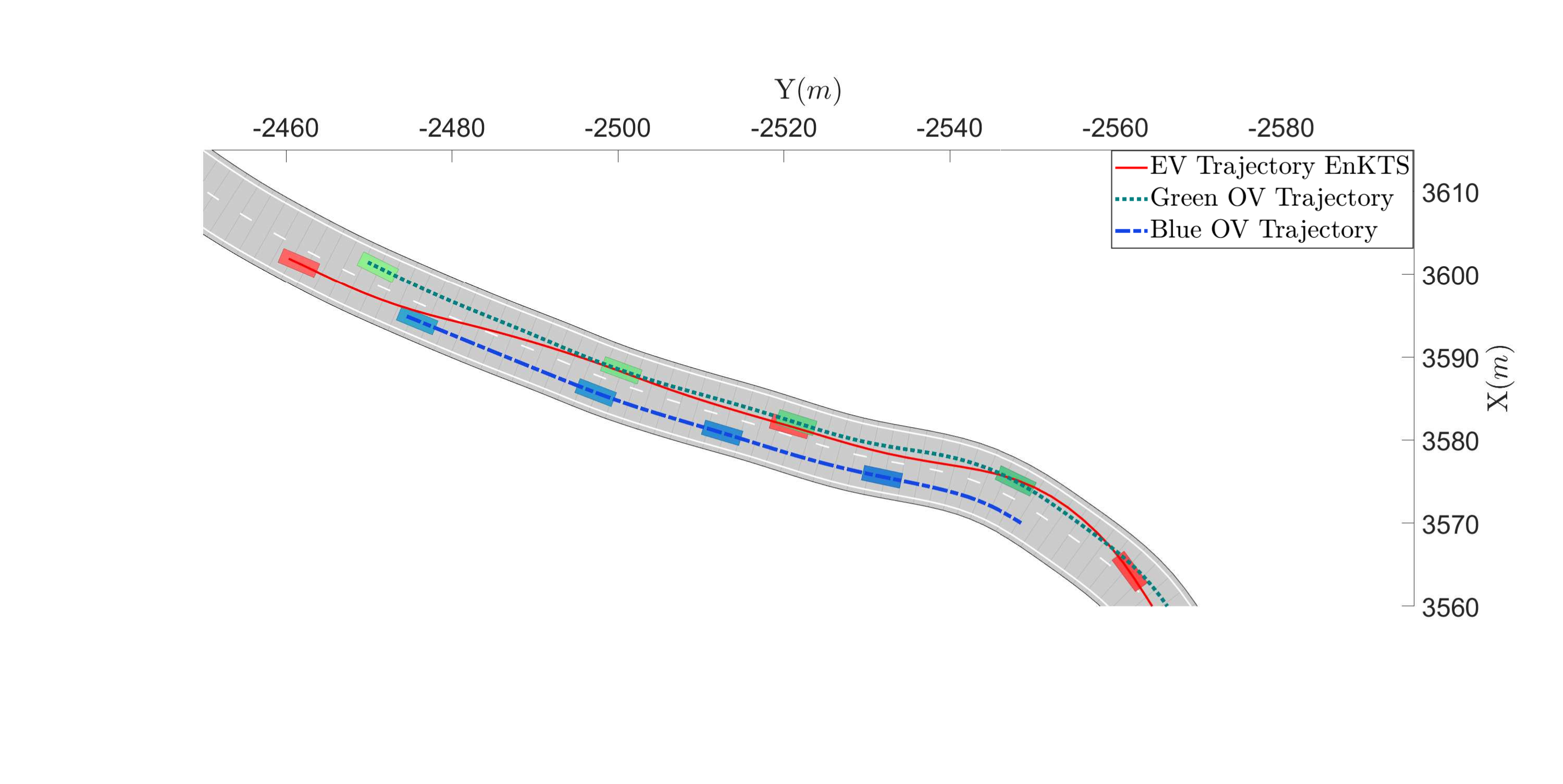}
        \subcaption{} \label{fig:Overtaking_Traj_EnKTS}
    \end{minipage}

    \vspace{-1mm}
    
    \caption{Overtaking scenario: trajectory generated by (a) EnKS and (b) \Kalmant.}
    \label{fig:Overtaking_Traj}
    
\end{figure*}
\begin{figure*}[t]
    \centering
    \subfloat[\centering ]{\label{fig:OverTakingcontrols-a}{\includegraphics[trim={5cm 9cm 3.5cm 9.5cm},clip=5cm, width=0.23\linewidth]{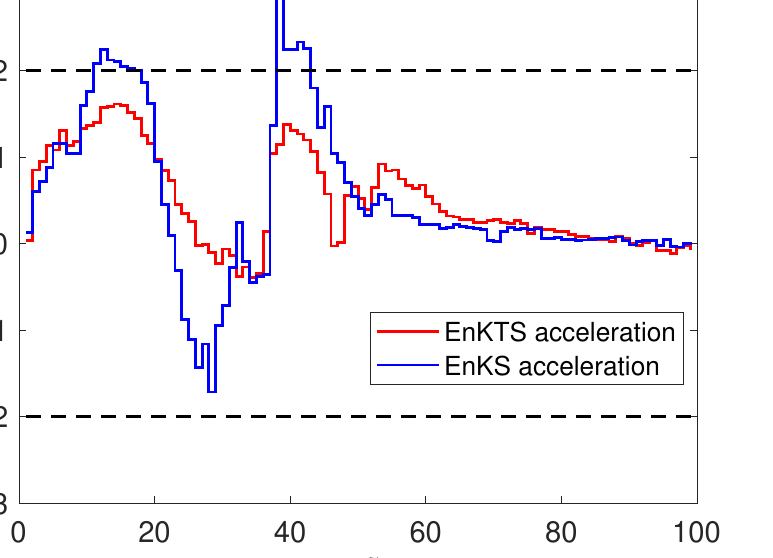}} }
    \subfloat[\centering ]{\label{fig:OverTakingcontrols-c}{\includegraphics[trim={5cm 9cm 3.5cm 9.5cm},clip=5cm, width=0.23\linewidth]{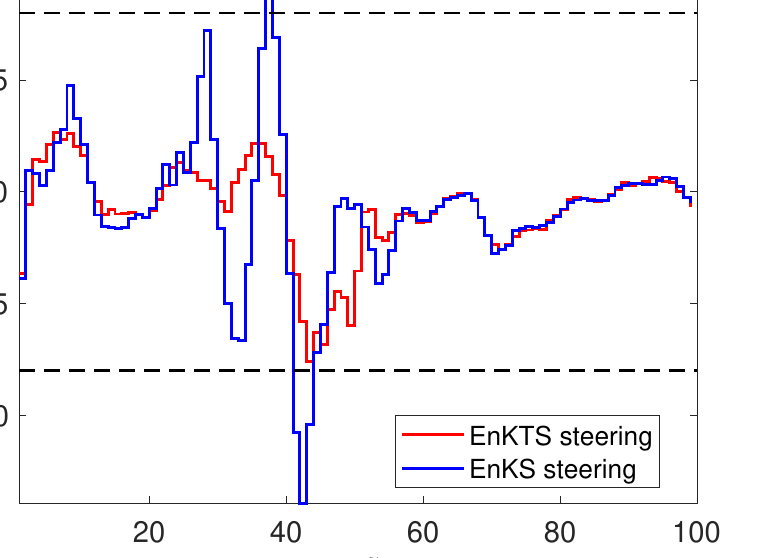}} }
    \subfloat[\centering ]{\label{fig:OverTakingcontrols-d}{\includegraphics[trim={4.8cm 9cm 3.4cm 9.5cm},clip=5cm, width=0.235\linewidth]{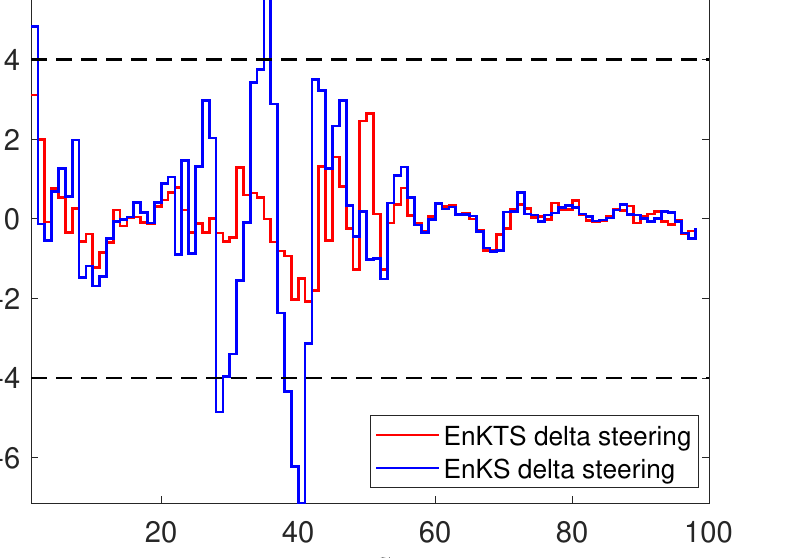}}}
    \subfloat[\centering ]{\label{fig:OverTaking-road}{\includegraphics[trim={4.9cm 9cm 3.4cm 9.5cm},clip=5cm, width=0.235\linewidth]{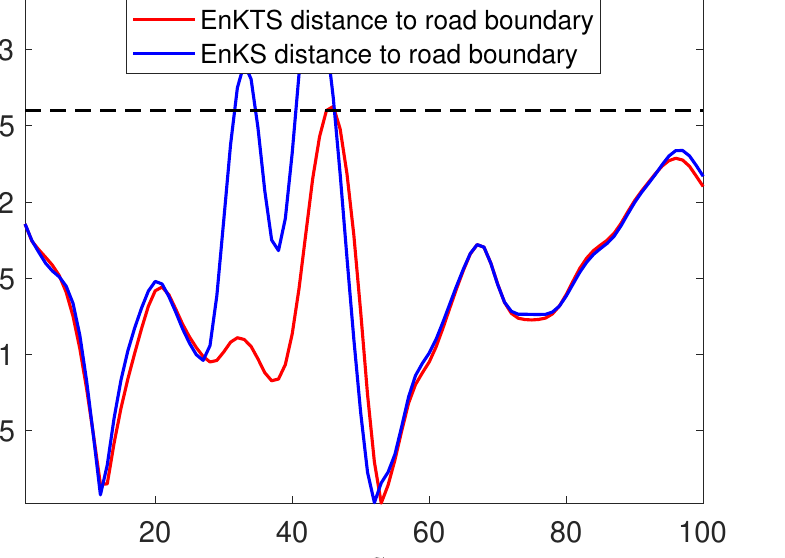}} }
    \caption{The EV's control profile and distance to road boundaries in the overtaking scenario. Comparison of \Kalmant and EnKS for (a) acceleration, (b) steering, (c) incremental steering, and (d) distance to the road boundary. Black lines show the constraints imposed on the variables and also the safety margin for the distance to road boundary.}
    \label{fig:Controls_overtaking}
\end{figure*}

\subsection{The Overtaking Scenario}
In this scenario, the EV and OVs travel on a curved, two-lane highway. As illustrated in Fig.~\ref{fig:Overtaking_Traj}, the EV (red) starts behind two slower OVs (green and blue). As shown in Fig.~\ref{fig:Controls_overtaking}, the EnKS violates the input constraints most of the time and crosses the road boundary twice. 
We observe that the EV leaves the road boundaries at time steps $38$ and $43$ during the simulation. This behavior occurs because the Gaussian distribution tends to limit exploration, failing to sample plans in low-probability regions far from the motion plan computed at earlier time steps.
In contrast, the \Kalmant algorithm consistently respects the constraints and remains within the road boundaries. Our observations indicate that EnKS only performs well under conditions with slacked barrier functions and less stringent inequality constraints. After time step $60$, when the EV has passed the OVs, the motion plans computed by both EnKS  and \Kalmant show the same performance. The reason is because there is no dynamic obstacle ahead of the EV which facilitates constraint satisfaction for EnKS. These findings underscore the effectiveness of the \Kalmant algorithm in generating safe motion plans due to leveraging heavy-tailed Student's-$t$ distribution, even in the presence of highly nonlinear dynamics, stringent constraints, and tight barrier functions. Next, we will consider another scenario and show that under a sudden environmental change, the performance of the EnKS deteriorates, highlighting the vulnerability of short-tailed distributions in ensuring safe motion planning.

\subsection{Emergency Brake Scenario}
\begin{figure*}[t]
    \centering
    \begin{subfigure}[b]{\textwidth}
        \centering
        \includegraphics[trim={2cm 7cm 2cm 8cm}, clip, width=\linewidth]{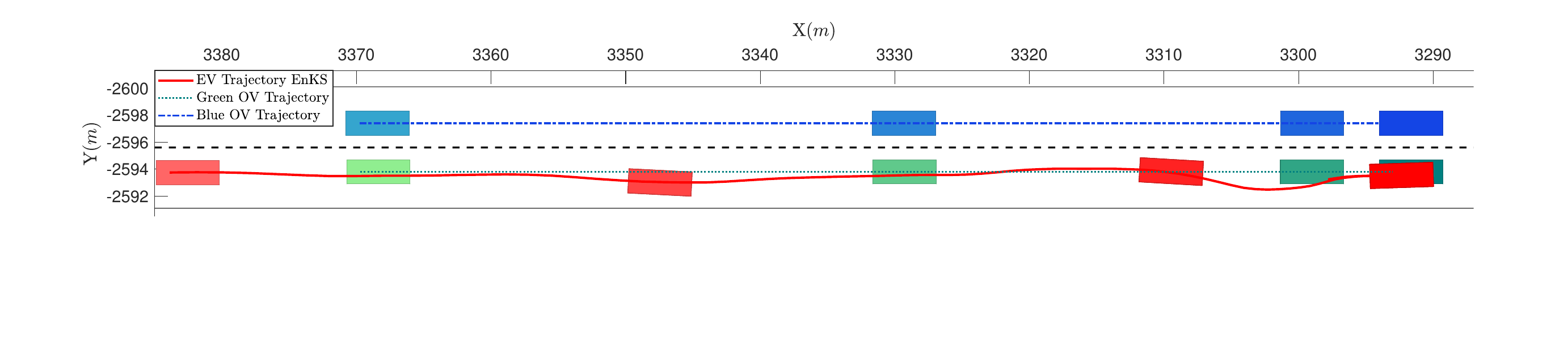}
        \vspace{-20mm}
        \subcaption{} \label{fig:EmergencyBrake_Traj_EnKS}
    \end{subfigure}

    \begin{subfigure}[b]{\textwidth}
        \centering
        \includegraphics[trim={2cm 7cm 2cm 7cm}, clip, width=\linewidth]{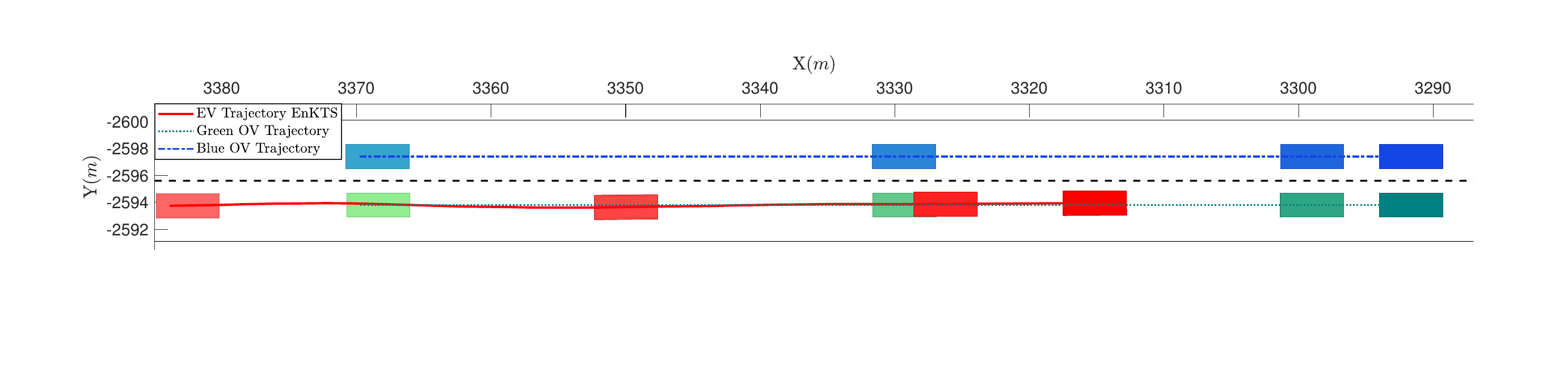}
        \vspace{-20mm}
        \subcaption{} \label{fig:EmergencyBrake_Traj_EnKTS}
    \end{subfigure}
    
    \caption{Emergency stop scenario: trajectory generated by (a) EnKS and (b) \Kalmant.}
    \label{fig:EmergencyBrake_Traj}
\end{figure*}

\begin{figure*}[t]
    \centering
    \subfloat[\centering ]{\label{fig:Emergencycontrols-a}{\includegraphics[trim={5cm 8.5cm 3.5cm 9cm},clip=5cm, width=0.23\linewidth]{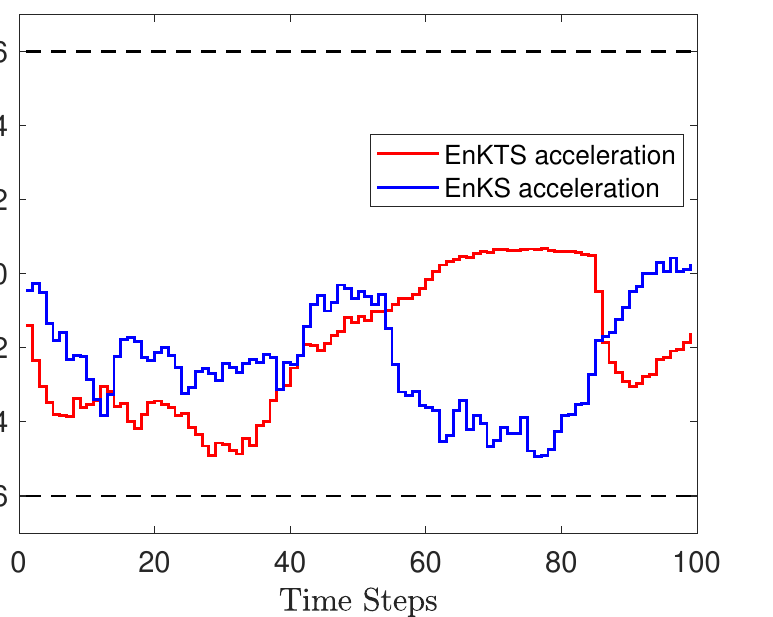}} }
    \subfloat[\centering ]{\label{fig:Emergencycontrols-b}{\includegraphics[trim={5cm 8.5cm 3.5cm 9cm},clip=5cm, width=0.23\linewidth]{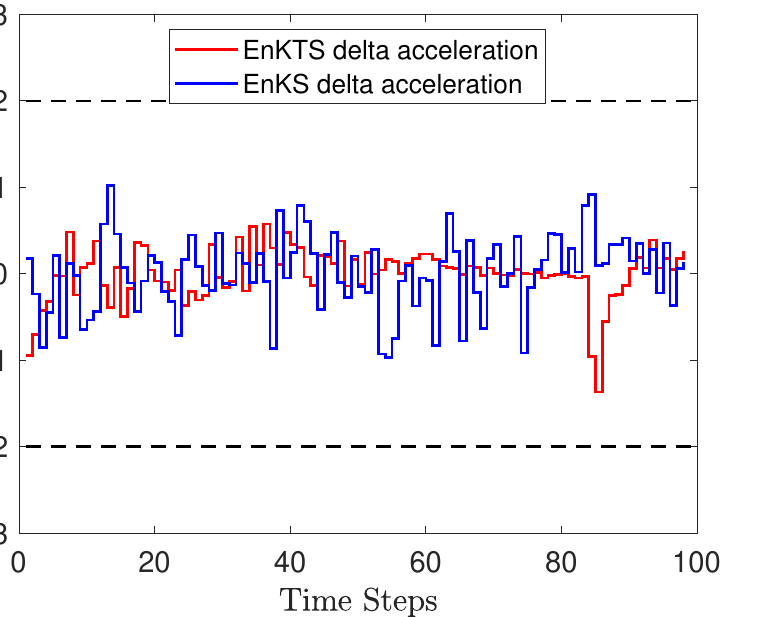}} }
    \subfloat[\centering ]{\label{fig:Emergencycontrols-c}{\includegraphics[trim={5cm 8.5cm 3.4cm 9cm},clip=5cm, width=0.23\linewidth]{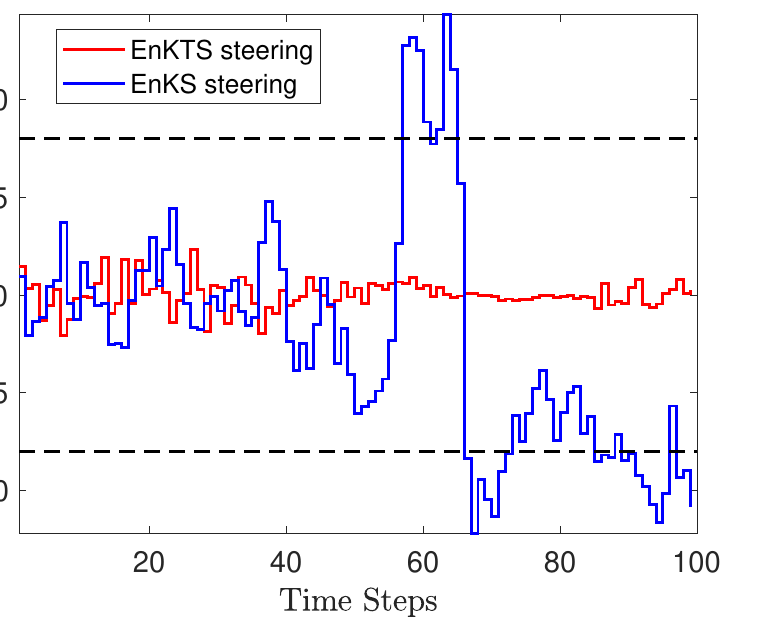}} }
    \subfloat[\centering ]{\label{fig:Emergencycontrols-d}{\includegraphics[trim={4.8cm 8.5cm 3.5cm 9cm},clip=5cm, width=0.23\linewidth]{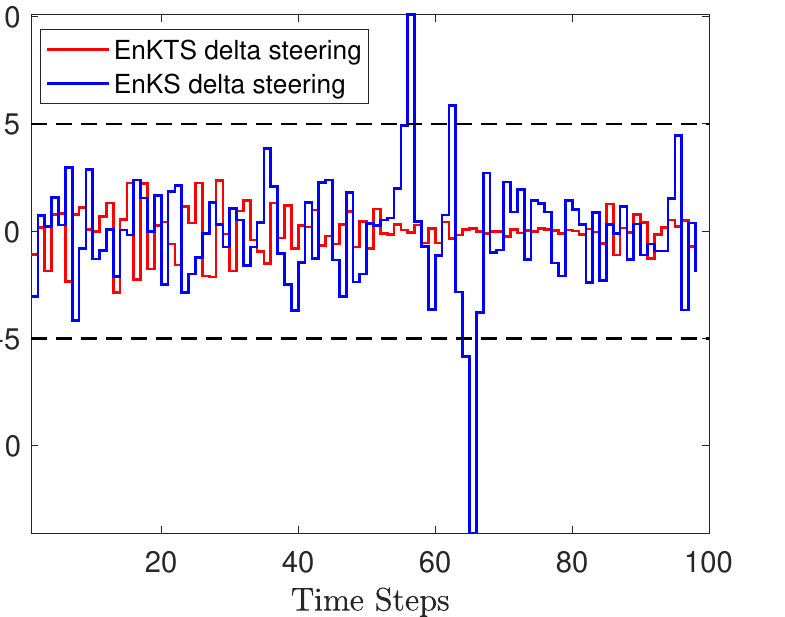}}}
    \vspace{-2mm}
    \caption{The control profile and constraint satisfaction by the EV in the emergency stop scenario. Comparison of \Kalmant and EnKS for acceleration (a,b) and steering (c,d) control profiles along with their incremental change. Black lines show the constraints imposed on the variables.}
    \label{fig:EmergencyControls}
\end{figure*}

\begin{figure*}[t]
    \centering
    \subfloat[\centering ]{{\includegraphics[trim={5cm 8.5cm 3.5cm 10cm},clip=5cm, width=0.23\linewidth]{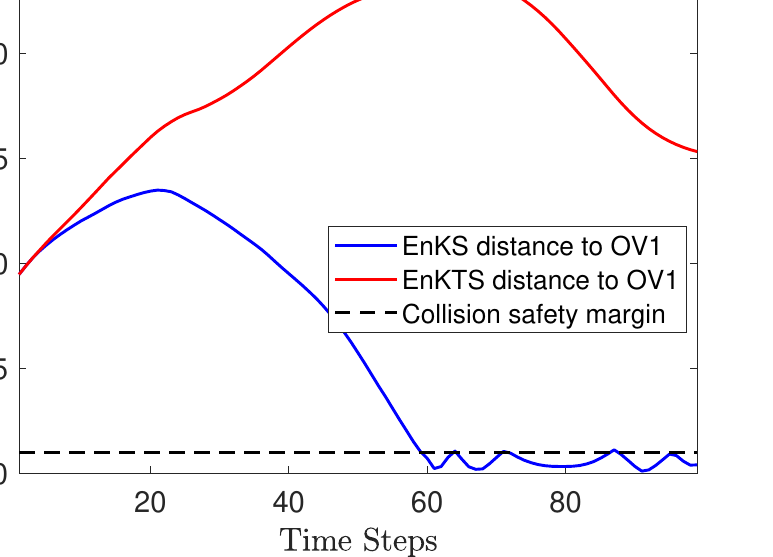}} }\label{fig:Emergency-OV1}
    \subfloat[\centering ]{\label{fig:Emergency-OV2}{\includegraphics[trim={5cm 8.5cm 3.5cm 10cm},clip=5cm, width=0.23\linewidth]{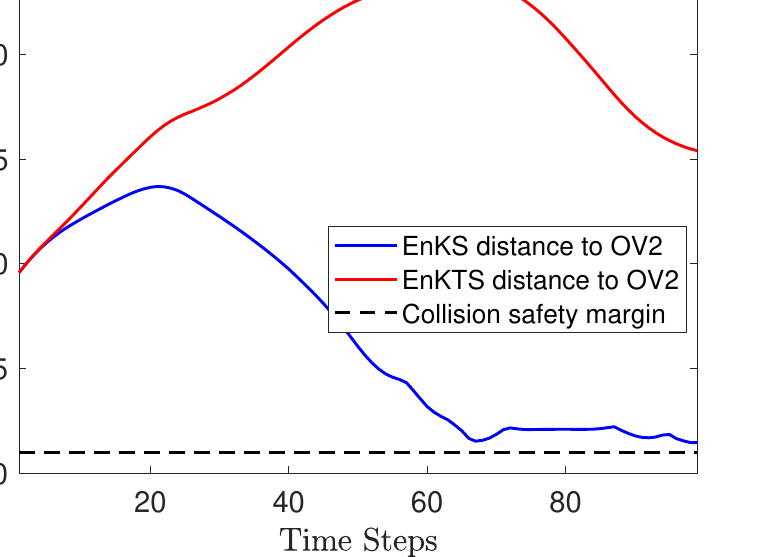}} }
    \subfloat[\centering ]{\label{fig:Emergency-road}{\includegraphics[trim={5cm 8.5cm 3.4cm 10cm},clip=5cm, width=0.23\linewidth]{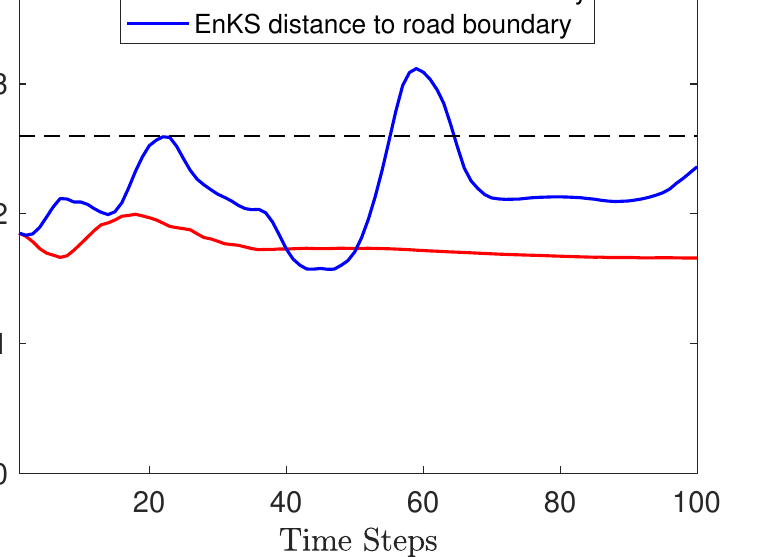}} }
    \subfloat[\centering ]{\label{fig:Emergency_speedref}{\includegraphics[trim={4.8cm 8.5cm 3.5cm 10cm},clip=5cm, width=0.23\linewidth]{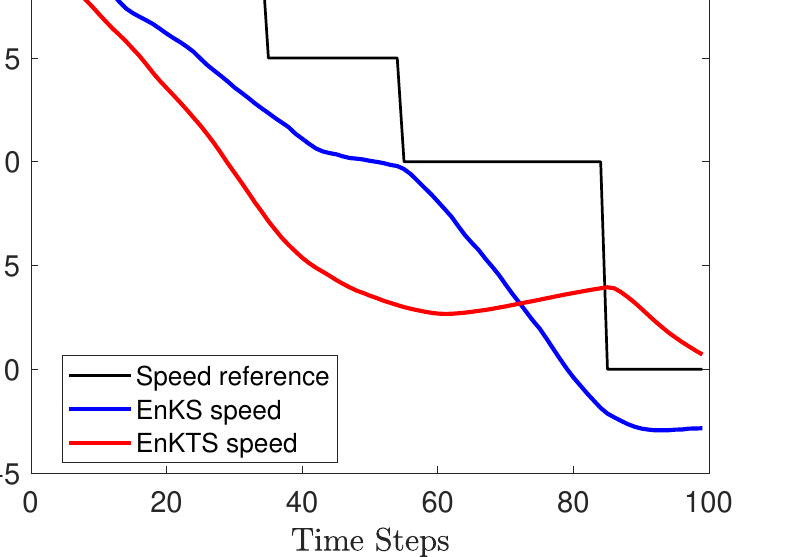}}}
    \vspace{-2mm}
    \caption{Comparison of distances to (a) OV1, (b) OV2, and (c) road boundary in the overtaking scenario with the black lines showing the safety margins. Reference speed value generated by the higher-module route planner and the tracking performance of both \Kalmant and EnKS is depicted in (d).}
    \label{fig:Controls}
    \vspace{-5mm}
\end{figure*}

Highway traffic accidents are often triggered by sudden and unpredictable events, such as rapid traffic build-up due to unexpected stops often trigger highway traffic accidents. In this scenario, we simulate a sudden traffic build-up by having the OVs come to a stop quickly. This requires the motion planner to react swiftly to safely decelerate to a complete stop., all while maintaining passenger comfort and preventing collisions with the OVs ahead. Fig.~\ref{fig:EmergencyBrake_Traj} shows the EV's motion plan using \Kalmant and EnKS. During the first three seconds, the high-level route planner is assumed to be unaware of the impending congestion, thus maintaining the reference speed unchanged as shown in Fig.~\ref{fig:Emergency_speedref}. The EV begins to decelerate, as depicted in Fig.~\ref{fig:Emergencycontrols-a},~\ref{fig:Emergencycontrols-b} but the magnitude of deceleration in \Kalmant is larger than that of EnKS. The EV controlled by EnKS crosses the road boundary during deceleration, requiring large steering adjustments with incremental corrections to return to the lane. At time step 50, the EV's speed under EnKS remains high because EnKS couldn't explore a broader decision space and resisted changing its current speed estimates. Meanwhile, the gap between the EV and the OV narrows, and although EnKS attempts to decelerate, it is insufficient. EnKS initiates a lane change by generating large steering inputs, but the EV crosses the road boundary again before returning to the lane. As the EV's speed still hasn't decreased enough, a collision with OV1 occurs at time step 60.In contrast, the EV controlled by \Kalmant operates smoothly, with minimal steering adjustments~\ref{fig:Emergencycontrols-c},~\ref{fig:Emergencycontrols-d}  to reduce speed, achieving collision-free motion planning~\ref{fig:Emergencycontrols-a},~\ref{fig:Emergencycontrols-b}. This is because \Kalmant has also explored lower probability regions where more rapid decelerations reside. At time step $80$, since the reference speed in Fig.~\ref{fig:Emergency_speedref} drops to zero, the motion planner decides to further decelerate the EV for a full stop, until the speed gets zero. These results show that heavy-tailed distributions such as the Student's-$t$ distribution reduce decision-making risks in safe motion planning by offering greater exploration capabilities. The heavy tails in the probability density function allow for more flexible updates to prior information when new measurement likelihoods are received, resulting in less resistance to adapting to changes.

\section{Conclusion} \label{sec:Conclusion}

Motion planning is a fundamental aspect of various robotic systems, where BIMP has proven to be a valuable approach. At the core of BIMP lies a probabilistic inference-driven search for motion plans based on planning objectives and constraints. However, despite its advantages, the performance of BIMP can be limited  by the use of short-tailed distributions, which may restrict its ability to explore less probable but high-quality plans. This limitation will become more critical for some complex problems, e.g., when highly nonlinear robot models are used. 
In this paper, we propose the use of heavy-tailed distributions to enhance the probabilistic search process. Specifically, we leverage   Student's-$t$ distributions and develop a Monte Carlo sampling-based single-pass smoothing approach to infer motion plans. This novel approach is also examined from the lens of NMPC. The simulation results demonstrate the effectiveness of this method, showing significant improvements in planning performance, constraint satisfaction, and sample efficiency. The proposed approach can find prospective use in motion planning for various other robots, and heavy-tailed distributions can be a useful way to facilitate other decision-making processes for robots.

\balance

\bibliographystyle{IEEEtran}
\bibliography{references}

\end{document}